\documentclass[journal]{IEEEtran}
\IEEEoverridecommandlockouts
\usepackage{cite}
\usepackage{amsmath,amssymb,amsfonts}
\usepackage{algorithm}
\usepackage{algorithmic}
\usepackage{graphicx}
\usepackage{textcomp}
\usepackage{authblk}
\usepackage{xcolor}

\def\BibTeX{{\rm B\kern-.05em{\sc i\kern-.025em b}\kern-.08em
    T\kern-.1667em\lower.7ex\hbox{E}\kern-.125emX}}
\begin{document}
\title{Deep Learning Based Intelligent Inter-Vehicle Distance Control for 6G Enabled Cooperative Autonomous Driving
\thanks{This work was supported by National Key R\&D Program of China (No.2018YFE0117500), the National Natural Science Foundation of China under Grant No. 62071092, the Science and Technology Program of Sichuan Province, China (No.2019YFH0007), and the EU H2020 Project COSAFE (MSCA-RISE-2018 under grant 824019).}
}

\author[1]{Xiaosha Chen}
\author[1,*]{Supeng Leng}
\author[2]{Jianhua He}
\author[1]{Longyu Zhou}
\affil[1]{School of Information \& Communication Engineering,
\authorcr{University of Electronic Science and Technology of China, Chengdu, China}}
\affil[2]{School of Computer Science and Electronic Engineering, \authorcr{University of Essex, UK}}
\affil[*]{\small{Corresponding author, email: spleng@uestc.edu.cn}}

\maketitle

\begin{abstract}
Research on the sixth generation cellular networks (6G) is gaining huge momentum to achieve ubiquitous wireless connectivity.
Connected autonomous driving (CAV) is a critical vertical envisioned for 6G, holding great potentials of improving road safety, road and energy efficiency.
However the stringent service requirements of CAV applications on reliability, latency and high speed communications
will present big challenges to 6G networks.
New channel access algorithms and intelligent control schemes for connected vehicles are needed for 6G supported CAV.
In this paper, we investigated 6G supported cooperative driving, which is an advanced driving mode through information sharing and driving coordination.
Firstly we quantify the delay upper bounds of 6G vehicle to vehicle (V2V) communications with hybrid communication and channel access technologies.
A deep learning neural network is developed and trained for fast computation of the delay bounds in real time operations. 
Then, an intelligent strategy is designed to control the inter-vehicle distance for cooperative autonomous driving. 
Furthermore, we propose a Markov Chain based algorithm to predict the parameters of the system states, 
and also a safe distance mapping method to enable smooth vehicular speed changes. 
The proposed algorithms are implemented in the AirSim autonomous driving platform. 
Simulation results show that the proposed algorithms are effective and robust with safe and stable cooperative autonomous driving, 
which greatly improve the road safety, capacity and efficiency. 
\end{abstract}


\section{Introduction}

With more than 1 million fatalities on the road globally and increasing global warming, there are urgent needs to reduce road accidents and energy consumption.
As more than 80\% road accidents were caused due to human errors, autonomous driving, which is boosted by advanced sensors and recent breakthroughs on deep learning technologies, is widely accepted as one of the most promising technologies to tackle road safety challenges. 
However, due to the inherent limitations of sensors for autonomous driving and the complicated driving conditions, 
autonomous driving can bring its own road safety problems, there is still a long way to go for full autonomous driving
and road safety. 
As a special form of Internet of Things (IoT), connected vehicles has been developed in parallel for more than a decade to solve the road safety and efficiency problems. More recently, connected autonomous vehicles (CAV) with cooperative sensing and driving \cite{Zhou}\cite{CAV} were proposed as a new paradigm for future transports \cite{Yang}\cite{SunG}.

Vehicle to Vehicle (V2V) communication plays a critical role for the cooperative driving. 
Due to the highly dynamic and complicated road environment, the V2V communication systems are expected to provide high speed, ultra-reliable and low latency communication services for CAVs, which may not be met by the 5G V2X \cite{LTE-C-V2X} \cite{5G-V2X}.  
Research on the sixth generation cellular networks (6G) is gaining huge momentum to achieve up to 1 Tera bits per second and ubiquitous wireless connectivity.
The candidate enabling technologies including Terahertz, space-terrestrial networking, reconfigurable intelligent surface and artificial intelligence \cite{terahertz} \cite{6GFeature}. CAV is a critical vertical envisioned for 6G.
However, new channel access algorithms and intelligent control schemes for connected vehicles needed to meet the stringent communication requirements of CAV applications. 
As safety is ultimately important for CAV applications, it is also necessary to fully model the V2V communication system performance and explore its impact on the CAV application safety, especially under the worst network conditions. Nevertheless, such tasks are very challenging. To the best of our knowledge, there is no reported work on the modelling and quantifying the performance guarantee of V2V communications with hybrid communication and channel access technologies.

In view of the above research challenges, we are motivated to investigate the communication performance of 6G V2V networks with hybrid communication and channel access schemes, and apply the findings to the intelligent control of cooperative driving. 
Cooperative driving is taken as a representative CAV driving mode, which is enabled through information sharing and driving coordination among vehicles.
Firstly we analytically model the communication delay upper bounds of the V2V networks by Stochastic Network Calculus (SNC), 
which is a popular mathematical tool to analyze the performance low bound of wireless communication systems. 
While the analytic model is effective and helpful, the high computation time prevents its use in real time network operation. 
To address this problem, a deep learning neural network is developed and trained for fast prediction of the delay bounds in real time operations. 
The analytical  results obtained from the SNC are used to create training samples for the deep learning neural network.
With the newly available V2V communication performance bounds, we propose an intelligent strategy for the control of inter-vehicle distance for cooperative
autonomous driving. 
Furthermore, we propose a Markov Chain based algorithm to predict the parameter and a safe  distance mapping method, which are utilized to enable smooth
vehicular speed changes \cite{phantomJam}.

Our contributions are summarized as follows.

\begin{itemize}
	\item We propose a SNC based theoretical tool set to derive the communication delay upper bounds for a 6G V2V network with hybrid communication and channel access schemes. Furthermore, a new analytical paradigm combining with the deep neural network and SNC theory is proposed for the real-time performance evaluation in a CAV environment.
	\item With the communication latency bounds, we propose an intelligent inter-vehicle distance control strategy for CAV. With the aid of the pre-trained deep neural network and our proposed on-line prediction method, each vehicle can determine its speed individually based on the states of other vehicles. Parameters prediction and distances mapping algorithms are designed to enable smooth speed changes, which can improve the efficiency of road traffic and energy consumption.
	\item We implement a comprehensive simulation platform which integrates AirSim based cooperative autonomous driving, Mathematica based SNC computing, communication and machine learning simulation. Extensive simulation experiments demonstrate that the proposed intelligent strategy works excellently in CAV scenarios. It is effective and robust  in terms of safety and stability with smooth speed changes.
\end{itemize}

The rest of this paper is organized as follows. 
Section \ref{related} introduces the related work of connected vehicles and autonomous driving. 
Section \ref{system} presents the system model and assumptions. 
Section \ref{analysis} shows the analysis of the proposed 6G-enabled cellular vehicular network. We present our proposed vehicular control algorithm in Section \ref{following}. 
Numerical simulation results and discussions are presented in section \ref{simulation}. 
Finally section \ref{conclusion} concludes the paper.

\section{Related Work}
\label{related}

Road accidents and traffic congestion have been global challenges for long time. Equipped with advanced sensors and deep learning technologies modern vehicles are becoming more autonomous and safer. However, due to the limitations of the sensors (such as limited sensing ranges and poor performance of cameras and radars in adverse weather conditions), the safety performance of these vehicles is still far from being satisfactory.  
On the other hand, connected vehicles is opening space for cooperative driving, which can be an excellent complement to provide a critical virtual sensor for autonomous vehicles. 

Vehicle to Everything (V2X) technologies have been proposed for communications among vehicles, between vehicles and infrastructure, pedestrians and networks \cite{LTE-C-V2X, 5G-V2X, 11bd}. 
In the latest 5G Release 16 specifications frozen in June 2020, support for enhanced V2X has been included, including advanced driving, extended sensors and vehicle platooning \cite{5G-V2X}. IEEE is also working on 802.11bd standard, which is an extension of 802.11p to provide advanced V2X support \cite{11bd}.
While the existing V2X technologies can provide basic connectivities for vehicles,
these technologies cannot meet the strict requirements on the communication latency and data rate from the advanced cooperative driving applications
and provide intelligent support to ensure the safety under very challenging communication and road conditions \cite{safety,Huang2017,Xiong2019,Qiao2018}.

As a promising technology for improving road efficiency and safety, 6G based V2V communications were studied by a few works. 
Zhifeng Yuan \emph{et. al} evaluated the efficiency of 6G full-duplex V2V channels in \cite{6GFull-Duplex}. 
Fengxiao Tang \emph{et. al} provided a survey on the application of machine learning  for 6G vehicular network. 
However, research on 6G technologies is still in an early stage. There is no report on the new design of channel access for the 6G V2V networks with hybrid communicaiton technologies. 

Some previous work studied the delay upper bound for different types of vehicular networks. In  \cite{DCF}, Jing Xie \emph{et. al} obtained the stochastic delay upper bound for the IEEE 802.11 DCF network, which formed the basis of many following works. The work \cite{DSRC} studied the end-to-end delay upper bound for the V2X network. However, they only consider the DSRC for the V2V network, without consideration of mmWave communication part. 
Guang Yang \emph{et. al} analyzed the multi-hop mmWave networks with full-duplex buffered relays in \cite{mmWave}, where Moment Generating Function (MGF) based upper bound for the service curve of the mmWave network was obtained. However, the analysis of the V2V delay upper bound in the previous works did not consider the hybrid access V2V network, which is consisted of different types of V2V communications links.

From the perspective of V2V based cooperative perception, Nunen \emph{et. al} \cite{Nunen} proposed an intended acceleration prediction based on V2V perception,
which demands sufficient time to ensure high performance and robustness in terms of string stability and driving safety. Nekoui \emph{et. al} \cite{Nekoui} demonstrated the V2V communication improves road traffic capacity by reducing driver perception-reaction time, which implies a high-speed compact platoon. Besides, Katsaros \emph{et. al} \cite{DSRC} analyzed the upper bound of the end-to-end delay for location-based routing in a hybrid vehicular network.

Kei Sakaguchi \emph{et. al} proposed a safe distance determination algorithm in mmWave V2V networks in \cite{Kei}. In \cite{Deng}, Deng Pan \emph{et. al} proposed a smart following strategy for autonomous vehicles. However, the vehicle following strategy in \cite{Deng} did not consider the communication performance in the vehicular network scenarios. Jeroen C. Zegers \emph{et. al} proposed an inter-vehicle distance control strategy based on the sensor data in \cite{VCDC} by extending the work in \cite{SCDC}. Nevertheless, the real-time sensor data based control strategy fluctuates in a relatively wide range, which reduces the road and energy efficiency. 

In summary,  the performance analysis and traffic control have been investigated separately in the previous research work. Besides, the existing network calculus based analytical methods for the delay upper bound have very high computational complexity. Therefore, it is difficult to apply these methods in on-line algorithms. 
The machine learning technology is a promising way to implement real-time SNC-based algorithms, which is one of the main ideas of this paper.

\section{System Model}
\label{system}

In this section, we introduce the system model and the basic assumptions. As roadside units (RSU) may not be always available in all scenarios, we consider the worst cases of the vehicle networks for cooperative driving, where vehicles communicate through distributed V2V communications. 
In the distributed V2V communication networks, vehicles compete to access the broadcast channels with conventional  for data transmission in a random manner,
as specified in the DSRC and 3GPP LTE-V2V  technologies. 
In this paper we will take exponential back-off protocols as the random access protocols 
for communication performance modelling and design of inter-vehicle distance control strategies.
It is noted that proposed theory model in the paper are applicable to modern random access protocols for 6G V2V networks as well.
Furthermore, to improve the communication reliability and data rate, we assume that vehicles are also equipped 6G radio access device, such as the visible light communication unit, on top of the traditional 5G radio access device.  

For the sake of analytical model tractability, hidden node and exposed node problems are not investigated as done in the most previous analytical work. 
For the infrastructure-free V2V communication at relative low radio frequency, all vehicles follow the random channel access for data transmissions. 
They are assumed to be within the communication range of each other. 
While for the 6G V2V communication at high frequency bands (such as mmWave and Terahertz bands), due to the line of sight communication limitation, we assume that the vehicles can only communicate to the vehicles immediately next to them and within the communication range. 
Without loss of generality, we will take mmWave communication as an example for 6G V2V communication at high frequency bands.
Random channel access is not applied for 6G mmWave communications 
as there is no contention exists with other vehicles for communication with beamforming technologies.
But the random channel access based V2V communications at the low frequency bands will be used for the direct communications between vehicles.  
To differentiate the random channel access communications at low frequency and the direct communication at high frequency bands,
we call these communications as cellular V2V and 6G communications, respectively.
The V2V networks with both low frequency band and 6G higher frequency band is called hybrid networks.
The protocols and topology of the hybrid network will be presented along with the explanation of some essential assumptions. 

\begin{figure}[tb]
\centering
\includegraphics[width=.5\textwidth]{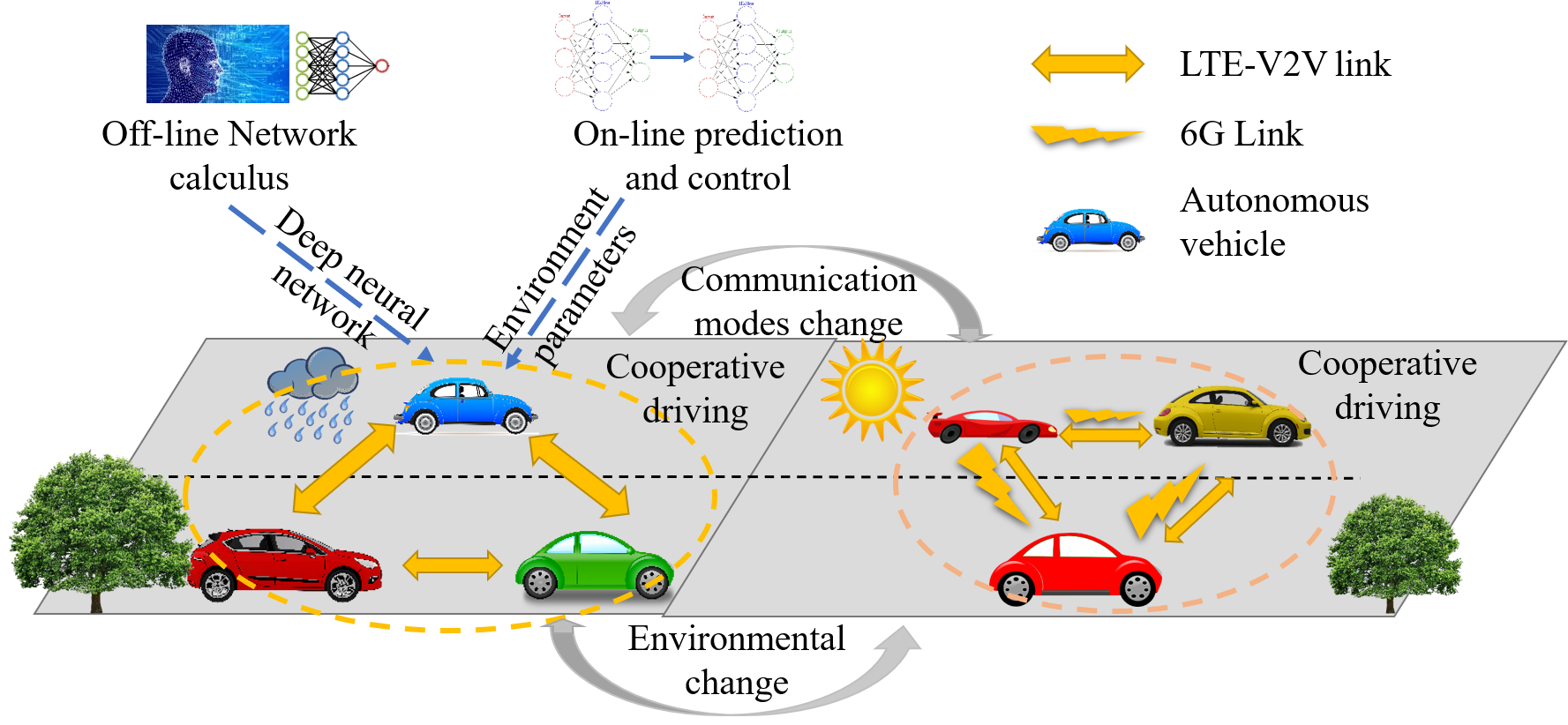}
\centering
\caption{A  scenario of 6G-supported cooperative connected vehicles.}
\label{fig:scenario}
\end{figure}

Fig. \ref{fig:scenario} shows an example 6G supported vehicular network under different weather conditions. In the good weather conditions, both cellular V2V and 6G communication channels can be established for short safety messages and large messages containing sensor data, separately. However, if the environment is not feasible to establish the 6G channels, the cellular V2V can also maintain the basic functions of the cooperative autonomous driving.
According to the characteristics of cellular V2V and 6G communications, it is reasonable to transmit short packets for wide range through cellular V2V communications
and large sensing data packets for close neighbours via 6G communications. If the 6G connectivity is not available, cellular V2V communication is needed to transfer part of the sensor data.
For the hybrid vehicular network, each vehicle has a message queue for cellular V2V communication and 6G communication separately. 
The vehicles have identical capacity of queues for the same communication type. The queue length of cellular V2V communication is denoted by $L$. 
Vehicles can establish 6G communication links only if the distance between them are shorter than a required threshold, such as the mmWave link in \cite{Park2013}. To simplify the performance analysis, we assume that the topology of the 6G communication network is a connected graph.
The 6G communication is assumed as a fluid-flow, discrete-time queuing system with a infinite buffer. 
Moreover, the longest flow in the 6G network is a flow with a start node $S$, $n-2$ full-duplex relays and a destination node $D$, where $n$ is the number of vehicles.

\subsection{Cellular V2V Communication}

For the cellular V2V communication, the system time is slotted. The communication follows the exponential back-off mechanism like IEEE 802.11p standard. In the back-off stage, we assume each communication process at a vehicle is in one of $M+1$ states, denoted by $B_0, B_1,\dots,B_M$ respectively. Each communication process also maintains a back-off counter and a content window. In the initial state $B_0$, the content window size is set to $\text{CW}_0=\text{CW}_{\min}=W$. The back-off counter is uniformly and randomly picked in the range $[0, \text{CW}_0]$. The back-off counter decreases by one if the channel is idle and will be frozen when the channel is busy. A vehicle tries to transmit the packet if its back-off counter equals to 0. If the transmit packet collides with other packet(s), the vehicle will transform to the next state and re-initialize the back-off counter in the range $[0, \text{CW}_i]$, where $i$ is the current state index.  If a conflict appears in the last state, the back-off process will start again from the initial state. The value of $\text{CW}_i$ can be obtained by
$\text{CW}_i=\min\left\{2^i W, 2^m W\right\},$
where $0<m\leq M$ is a parameter of the protocol to control the maximum content window size.

Note that for data broadcasting, messages are usually not retransmitted due to the difficulty of sending back ACK or NACK packets. And for many safety applications, periodic broadcast messages will need to carry the latest vehicle status information (such as location, speed and driving direction etc), retransmitted messages with excessive delay could be outdated and not useful for the safety applications. However, the back-off mechanism is still applied to the transmission of all packets for their first transmission attempt, which is important to control channel congestion, especially for distributed V2V communications.

\subsection{6G  Communication}

As a typical example of 6G communication links, we consider mmWave link in this section. As other types of 6G communication channels mainly differ on the channel model, so the analysis will be similar. We assume a homogeneous mmWave communication environment, where all the mmWave links follow the same fading distribution. 
Given the distance between the nodes, a general fading factor $g$ in dB is given by \cite{mmWaveGain,mmWaveGain2}.

\begin{equation}
\label{eq:mmWaveLink}
g[\text{dB}] = -(\alpha +10\beta \log_{10}(l)+\xi),
\end{equation}

\noindent where $\alpha$ and $\beta$ are the least square fits of floating intercept and slope of the best fit, and $\xi\sim \mathcal{N}(0, v^2)$, a log-normal shadowing effect with variance $v^2$. The parameters $\alpha$ and $\beta$ are dependent on the communication environment.

Furthermore, as the mmWave channels are full-duplex, self-interference needs to be considered. This paper adopts a widely used method to model the self-interference, which uses a coefficient $0\leq \mu \leq 1$ to characterize the coupling between the transmitter and receiver. The interference plus noise ratio (SINR) can be denoted by $\gamma$, which is computed by $\gamma = \kappa\cdot\omega\cdot g,$ where

\begin{equation}
\label{eq:SNR}
\omega = \left\{
\begin{array}{ll}
\displaystyle\frac{\text{SNR}}{1+\mu \text{SNR}} & i\in \{1,2,\dots, n-1\} \\
\text{SNR} & i=n
\end{array}
\right. ,
\end{equation}

\noindent and $\text{SNR}=\frac{P}{N_0}$ denotes the transmitted signal-to-noise ratio (SNR), where $P$ and $N_0$ are the transmitted power and background noise, respectively.

\section{Communication Delay Upper Bounds}
\label{analysis}

To determine the inter-vehicle safe distance, the communication delay upper bound needs to be computed. In this section we will analyze the stochastic delay upper bound of hybrid V2V  networks as introduced in section \ref{system}. 
We first show the performance analysis of the V2V network with only cellular V2V communication or 6G mmWave communication, respectively. And then we present the analysis of the proposed hybrid network with SNC theory. All queues studied in this paper are First-In-First-Out (FIFO) queues.

\subsection{Performance of Cellular V2V Communications}

To analyse the performance of the V2V network with cellular V2V communication, the collision probability will be analyzed first. According to \cite{collisionProb}, we have the following formula for the collision probability, which is denoted by $p_c$:

\begin{equation}
\label{eq:collisionProb}
\begin{array}{c}
\displaystyle p_c:=1-e^{-(n-1) p_a}\\
\displaystyle p_a:=\frac{\sum_{k=0}^M p_c^k}{\sum_{k=0}^M \mu_k p_c^k}
\end{array}
,
\end{equation}

\noindent where $p_c$ and $p_a$ is the probability that a collision happens in a time slot and a vehicle tries to transmit in a time slot, respectively. $\mu_k$ is the average size of the content window size of backoff stage $k$. The collision probability $p_c$ can be obtained by solve the non-linear equation system in Eq. \eqref{eq:collisionProb}, which is very fast by the numerical method.

We can calculate the mean service time for the random access channel by \cite{DSRC}

\begin{equation}
\overline{t}_{\mathrm{serv}}=\sum_{j=0}^M p_c^j \overline{t}_j,
\end{equation}

\noindent where $\overline{t}_j$ is the mean time that a node stays at back-off stage $j$, which can be calculated by

\begin{equation}
\overline{t}_j=\mu_j t_B+t_{T X},
\end{equation}

\noindent where $t_B$ is the average length of a back-off slot, $t_{TX}$ is the average length of a transmission slot. we can obtain $t_B$ by

\begin{equation}
t_B = (1-p_a)^n + n\cdot p_a(1-p_a)^{n-1}t_{T X}+p_c t_{C},
\end{equation}

\noindent where $t_C$ is the length of the collision back-off slot.

For the FIFO queue, the average waiting time can be obtained by the basic queue theory. We assume that the queue is a G/M/1 queue and according to the Pollaczek-Khinchin formula, the mean waiting time is

\begin{equation}
t_q = \rho + \frac{\rho^2 +\lambda^2 \text{Var}(S)}{2(1-\rho)},
\end{equation}

\noindent where $\lambda$ is the arrival rate for the Poisson process, $\rho=\lambda \overline{t}_{\mathrm{serv}}$ is the utilization, $\text{Var}(S)$ is the variance of the service time distribution $S$.

In terms of the Concatenation Property of SNC, and according to the work in \cite{DCF}, we can express the service curve of a node in cellular V2V network when $0 \leq y<1-q$ with

\begin{equation}
\label{eq:DSRCServ}
\beta(t) = \overline{t}_{\text{serv}}\lambda t \otimes t_q \lambda t
\end{equation}

\begin{equation}
\label{eq:DSRCBound}
g_t(x) = \left\{\left(\frac{q}{y}\right)^y\left(\frac{1-q}{1-y}\right)^{1-y}\right\}^L
\end{equation}

\noindent where

\begin{equation}
\arraycolsep=1.4pt\def\arraystretch{2.2}
\begin{array}{c}
\displaystyle q=\frac{\bar{t}_{\text{serv}}+t_q-t_s}{M t_C + L \bar{t}_{\text{serv}}+\mathcal{B}t_s}, \\
\displaystyle y=\frac{x-L t_s}{L(M t_C+L \bar{t}_{\text{serv}}+\mathcal{B}t_s)}
\end{array},
\end{equation}

\noindent where $\mathcal{B}=\sum_{k=0}^M(CW_k-1)$, $t_s$ is the average time that the channel is busy due to the successful transmission. $\otimes$ is the min-plus convolution in network calculus, which is defined by $(\alpha \otimes \beta)(t)=\inf _{0 \leq \tau \leq t}\{\alpha(\tau)+\beta(t-\tau)\}.$
Based on the SNC theory, the service process $S(t)$ of a node can be expressed by the stochastic service curve $\beta(t)$ bounded by $g_t(x)$, which is wrote as $S \sim_{\mathrm{sc}}\left\langle g_{t}, \beta\right\rangle$, and is defined as $P\left\{\sup _{0 \leq s \leq t}\left[A \otimes \beta(s)-A^{*}(s)\right]>x\right\} \leq g_{t}(x),$ where $A(t)$ and $A^*(t)$ is the arrival and departure process, respectively.

\subsection{Performance of 6G Communication Networks}

To simplify the analysis of the delay with 6G mmWave communication, which is assumed to be mmWave links as an typical example, we assume that all of the background noises for the links have the same power, which is called homogeneous mmWave network. As for the arrive process of the mmWave network, we assume that it is a $(\rho, \theta)$-upper bounded stochastic arrival process, which is defined as

\begin{equation}
M_A(\theta, s, t) \leq e^{\theta(\rho(\theta)(t-s)+\sigma(\theta))} \triangleq e^{\theta \sigma(\theta)}\left(p_{a}(\theta)\right)^{t-s},
\end{equation}

\noindent where $M_A(\theta, s, t)$ is the moment of the arrival process in time period $[s,t)$ with respect to $\theta$ and $p_a(\theta)=e^{\theta\rho(\theta)}$.

According to the work in \cite{mmWave}, the moment generating function of the service process in $n$-hop mmWave network has an upper bound, which is

\begin{equation}
\label{eq:mmWaveServ}
\mathrm{M}(\theta, s, t) \leq \frac{e^{\theta \sigma(\theta)}}{p_{a}^{s-t}(\theta)} \cdot \mathcal{G}_{\tau, n-1}\left(p_{a}(\theta) \hat{q}(-\theta)\right)
\end{equation}

\noindent for any $\theta>0$, where $\tau\triangleq \max(s-t,0)$ and $p_a(\theta)\hat{q}(-\theta)<1$. Moreover, $\hat{q}(-\theta)$ is defined the exception of $(1+\xi)^{-\theta}$, which can be calculated by

\begin{equation}
\begin{array}{l}
\hat{q}(-\theta)=\mathbb{E}\left[(1+\xi)^{-\theta}\right]\approx\\
\displaystyle\min _{u \geq 0}\left\{\left(1+\delta N_{\delta}(u)\right)^{-\theta}+\sum_{i=1}^{N_{\delta}(u)} a_{\theta, \delta}(i) F_{\xi}(i \delta)\right\}
\end{array}
\end{equation}

\noindent where $F_{\xi}(x)$ is the PDF of random variable $\xi$, $N_{\delta}(u)=\left\lfloor\frac{u}{\delta}\right\rfloor$ and $a_{\theta, \delta}(i)=(1+(i-1) \delta)^{-\theta}-(1+i \delta)^{-\theta}$, $\delta$ is a parameter, less $\delta$ means higher accuracy.

Moreover, in Eq. \eqref{eq:mmWaveServ}, $\mathcal{G}_{\tau, n}$ is defined as

\begin{equation}
\mathcal{G}_{\tau, n}(x) \triangleq \min \left\{\mathcal{G}_{1}(x), \mathcal{G}_{2}(x)\right\}
\end{equation}

\noindent where

\begin{equation}
\displaystyle \mathcal{G}_{1}(x)=\frac{\min \left(1, x^{\tau}\left(\begin{array}{c}{n+\tau} \\ {n}\end{array}\right)\right)}{(1-x)^{n+1}},
\end{equation}

\begin{equation}
\mathcal{G}_{2}(x)=\frac{1}{(1-x)^{n+1}}-\left(\begin{array}{l}{n+\tau} \\ {n+1}\end{array}\right) x^{\tau-1}
\end{equation}

After we obtain the moment generating function bound, we can get a probability upper bound by Chernoff Bound, which is

\begin{equation}
P\{X \leq x\} \leq e^{-\theta x} E e^{\theta X}=e^{\theta x} M_{X}(-\theta).
\end{equation}

Consequently, the service amount of the mmWave network can be written as

\begin{equation}
\label{eq:probBound}
P\{X\leq x\}\leq \frac{e^{\theta (\sigma(\theta)+x)}}{p_{a}^{s-t}(\theta)} \cdot \mathcal{G}_{\tau, n-1}\left(p_{a}(\theta) \hat{q}(-\theta)\right).
\end{equation}

Replace $x$ with $\beta(t-s)-x$ in Eq. \eqref{eq:probBound}, we can get

\begin{equation}
\begin{array}{l}
\displaystyle P\{X\leq \beta(t-s)-x\}\leq \\
\displaystyle \frac{e^{\theta (\sigma(\theta)+\beta(t-s)-x)}}{p_{a}^{s-t}(\theta)} \cdot \mathcal{G}_{\tau, n-1}\left(p_{a}(\theta) \hat{q}(-\theta)\right),
\end{array}
\end{equation}

Let $\beta(t-s)=\frac1\theta \ln p_a^{s-t}(\theta)=(t-s)\rho$, we can get the service curve for the mmWave network:

\begin{equation}
\label{eq:mmWave}
\begin{array}{l}
\displaystyle P\{X\leq \rho(t-s)-x\}\leq\\
\displaystyle e^{\theta (\sigma(\theta)-x)} \cdot \mathcal{G}_{\tau, n-1}\left(p_{a}(\theta) \hat{q}(-\theta)\right)\\
\end{array}
\end{equation}

From Eq. \eqref{eq:mmWave} and \cite{book}, we can find that the mmWave network provide a weak stochastic service curve for the flows.

\subsection{Delay Upper Bounds}

After obtaining the service curve, we can get the delay upper bound. For the arrival curve $\alpha(t)$ with bounding function $f(x)$ and the service curve $\beta(t)$ with bounding function $g(x)$, the virtual delay $d(t)$,  which is defined as

\begin{equation}
d(t)\triangleq \inf \left\{\tau: A(t) \leq A^{*}(t+\tau)\right\},
\end{equation}

\noindent can be calculated by

\begin{equation}
\label{eq:theorm}
P\{d(t)>h(\alpha(t)+x, \beta(t))\} \leq(f \otimes g)(x).
\end{equation}

For the parallel service curve 1 and 2, which is given by

\begin{equation}
\arraycolsep=1.4pt\def\arraystretch{1.5}
\begin{array}{l}
\displaystyle P\{X\leq \beta_1(t-s)-x\}\leq f_1(x)\\
\displaystyle P\{X\leq\beta_2(t-s)-x\}\leq f_2(x),
\end{array}
\end{equation}

\noindent where $f_1(x)$ and $f_2(x)$ are the two bounding functions of the parallel service curve, respectively. The total service curve can be obtained by

\begin{equation}
\label{eq:parallelPrim}
P\{X\leq \beta_1(t-s)+\beta_2(t-s)-2x\}\leq f_1 * f_2 (x)
\end{equation}

\noindent where $*$ is the Stieltjes convolution, which is defined as

\begin{equation}
(a * b)(x)\triangleq\int_{-\infty}^{\infty} a(x-y) d b(y).
\end{equation}

Replacing $2x$ with $y$ in Eq. \eqref{eq:parallelPrim}, we can obtain

\begin{equation}
\label{eq:parallel}
P\{X\leq \beta_1(t-s)+\beta_2(t-s)-y\}\leq f_1 * f_2 \left(\frac{y}{2}\right).
\end{equation}

From Eq. \eqref{eq:DSRCServ}, \eqref{eq:DSRCBound}, \eqref{eq:mmWave} and \eqref{eq:parallel}, we can get the stochastic service curve bound for the hybrid V2V network, which is

\begin{equation}
\label{eq:hybridBound}
\begin{array}{l}
P\{X\leq \bar{t}_{\mathrm{scrv}} \lambda t \otimes t_{q} \lambda t+\rho(t-s)-x\}\leq\\
\displaystyle-\frac{\theta}{2} \mathcal{G}_{\tau, n-1}\left(p_{a}(\theta) \hat{q}(-\theta)\right)\cdot\\
\displaystyle \int_{-\infty}^{\infty} \left\{\left(\frac{q}{y'}\right)^{y'}\left(\frac{1-q}{1-y'}\right)^{1-y'}\right\}^L  e^{\theta(\sigma(\theta)-x/2)} \mathrm{d}x,
\end{array}
\end{equation}

\noindent where

\begin{equation}
\displaystyle y'=\frac{x/2-L t_s}{L(M t_C+L \bar{t}_{\text{serv}}+\mathcal{B}t_s)}.
\end{equation}

With Eq. \eqref{eq:theorm} and \eqref{eq:hybridBound}, and according to the aggregation property of the arrival process in \cite{book}, we can obtain the stochastic delay upper bound. Moreover, with the same method, the delays of the pure cellular V2V and pure 6G vehicular network can also be obtained.

\section{Distance Control Strategy for Cooperative Connected Driving}
\label{following}

Based on the delay upper bound, we can develop intelligent  inter-vehicle distance control strategy for connected vehicles on one lane as shown in Fig. \ref{fig:scenario}. The proposed strategy needs to satisfy three requirements on safety, traffic efficiency and energy efficiency. 
Because of the high complexity of the delay computation, we use first devise a machine learning based approach to make the algorithm suitable for real time operation. 
We will split the control strategy into two parts, the calculation of safe distance and decision making.

\subsection{Parameter Prediction}

To improve the safety level of the traffic and improve the on-road experiment, the communication performance of the vehicular network needs to be predicted. In this section, we will present a simple but efficient and accurate neural network based prediction algorithm.

Let $X$ be a discrete random process, and there are $X_n$ states that the process $X$ can be. Moreover, the time space is also discrete. At time $t$, $X$ generate a value $x_t$, and the value of $x_t$ depends on the previous $n$ values, in other words, $X$ is a $n$-order Markov process, which means

\begin{equation}
\begin{array}{l}
P(x_t|x_{t-1},x_{t-2},\dots,x_0)=\\
P(x_t|x_{t-1},x_{t-2},\dots,x_{t-n}).
\end{array}
\label{eq:transProb}
\end{equation}

If the values of a parameter are continuous, such as the distance of the vehicles, we can discretize the values by dividing the real number into different levels. To learn the transition probability in equation \eqref{eq:transProb}, the training data needs to be gathered first. With enough training data, the transition frequency can be calculated by

\begin{equation}
\hat{P}(\hat{x}_t|\hat{x}_{t-1},\dots,\hat{x}_{t-n})=\frac{C(\hat{x},\hat{x}_{t-1},\dots,\hat{x}_{t-n})}{C(\hat{x}_{t-1},\dots,\hat{x}_{t-2})},
\label{eq:frequency}
\end{equation}

\noindent where $\hat{P}(\hat{x}_t|\hat{x}_{t-1},\dots,\hat{x}_{t-n})$ means the frequency of $\hat{x}_t$ when the previous $n$ values are $\hat{x}_{t-1},\dots,\hat{x}_{t-n}$, respectively. $C(x_1, \dots, x_n)$ is the number of sequence $x_1,\dots, x_n$ in the gathered data.

Learning the transition probability is essentially a data fitting problem. As the number of possible combinations of the $n$ values is too high in most cases, we apply deep neural network method to predict the transition probability. The inputs of the neural network are the $n$ previous values, respectively and the outputs of the neural network are the probability of the $X_n$ states.
However, if we directly use the results of equation \eqref{eq:frequency} to train the neural network, the deep neural network can be easily over-fitted as the frequency is not accurate when the data size is small. We formulate a confidence factor to avoid this problem. A confidence factor of sequence $x_1,\dots,x_n$ is defined as

\begin{equation}
F(x_1,\dots,x_n)=1-e^{\zeta\cdot C(x_1,\dots,x_n)},
\label{eq:confidenceFactor}
\end{equation}

\noindent where $\zeta$ is a negative real number which can improve the accuracy. According to equation \eqref{eq:confidenceFactor}, the frequencies of sequence $x_1,\dots,x_n$ need to be updated according to the following equation,

\begin{equation}
\hat{P}_u(\hat{x}|x_n,\dots,x_1)=\left\{
\begin{array}{ll}
\displaystyle \hat{P} - \frac{F(x_1,\dots,x_n)}{Z(x_1,\dots,x_n)}, & \hat{P}>0\\
\displaystyle \frac{F(x_1,\dots,x_n)}{Z(x_1,\dots,x_n)}, & \hat{P}=0,
\end{array}
\right.
\label{eq:frequencyUpdate}
\end{equation}

\noindent where $\hat{P}=\hat{P}(\hat{x}|x_n,\dots,x_1)$ and $Z(x_1,\dots,x_n)$ is the number of states that the frequency equals to 0 when the previous values are $x_1,\dots,x_n$ in the state space.

In this paper, we use a deep neural network structure with 4 layers, all the nodes in the neural networks being ramp nodes. 
The whole process of the proposed prediction algorithm is presented in Algorithm \ref{alg:prediction}.

\begin{algorithm}[t] 
\caption{Parameter Prediction Algorithm}

\begin{algorithmic}[1]
 \STATE \textit{Initialization}: If the values of parameter are continuous, discretize them into different levels.
 \STATE \textit{Initialization}: Initialize the neural network with random values set for the parameters.
 \WHILE {True}
  \STATE Gather $n_s$ sequence data with length $n+1$.
  \STATE Calculate the frequencies with equation \eqref{eq:frequency}.
  \STATE Update the frequencies with equation \eqref{eq:frequencyUpdate}.
  \STATE Train the neural network off-line with the obtained results.
  \STATE Predict the next values on-line of the parameters with the trained neural network.
  \ENDWHILE
  
 \end{algorithmic} 
\label{alg:prediction} 
\end{algorithm}

\subsection{Distance and Speed Control}

The core task of the vehicular distance control strategy is to decide the distance between the vehicles and the speed of each vehicle. In this section, we will present how to determine the safe distance according to the results in section \ref{analysis} and how to avoid the high computation complexity of the network calculus by using deep neural network.

The safe distance means if the preceding vehicle starts to bake at its highest acceleration, then after receiving the control message, the current vehicle behind the preceding vehicle starts to bake, too. Any collisions need to be avoided in this process. According to this description, we can formulate the safe distance as following equation,

\begin{equation}
S_{i,i+1}=\frac{v_i^2}{2a_i}+D_{i,i+1}\cdot v_i-\frac{v_{i+1}^2}{2a_{i+1}},
\label{eq:safetyDistance}
\end{equation}

\noindent where $v_i, v_{i+1}$ is the speed of current vehicle and preceding vehicle, respectively. $a_i, a_{i+1}$ is the maximum acceleration of current and preceding vehicle, respectively. $D_{i,i+1}$ is the delay upper bound calculated by network calculus.

However, different from the wired link, the wireless communication is a stochastic process, it is not possible to provide an exactly delay upper bound. As we can see in section \ref{analysis}, the delay upper bound of the vehicular network is a probability distribution. Consequently, according to the dangerous level of different roads, the time-out probability upper bound $p$ needs to be provided. For example, on a road that traffic accident happens frequently, we need the time-out probability cannot be higher than 0.01, which means $p=0.01$. On the other hand, if the road is safer, the $p$ can be lower, such as $0.1$.

Nevertheless, calculating the delay upper bound given $p$ according to the network calculus needs dramatic computation resources and will lead to high computation delay. The safe distance decision process needs to be finished at a ultra-fast speed to guarantee the road safety and make the algorithm on-line. Our solution calculates the delays of different parameters and time-out probabilities off-line, then train a network with given data. The input of the neural network is the communication parameters and the probability, the output is the delay upper bound. Because the calculation is environment independent, this calculation even can be finished at the cloud and the results can be downloaded by vehicles before they run on roads.

If we adjust the distance between vehicles according to the safe distance, because of the high dynamics of the environment, the distance will change at each time, which will make the vehicles change their speeds frequently. This will cause vehicular energy wasting and reduce the on-road experience. Furthermore, the frequent acceleration and deceleration will cause Phantom Traffic Jams, which will heavily reduce the road traffic efficiency. To reduce the speed change, we adopt two methods: parameter prediction and safe distance mapping.

In the safe distance process, we consider not only the current safe distance, but also the longest safe distance in the next $T_n$ time by parameter prediction. Vehicles will predict the network parameters individually with Algorithm \ref{alg:prediction}, then calculate the delay upper bounds and the safe distances with the trained neural network by equation \eqref{eq:safetyDistance}, and adjust the speed according to the worst case.

\begin{figure}[htbp]
\centerline{\includegraphics[width=0.4\textwidth]{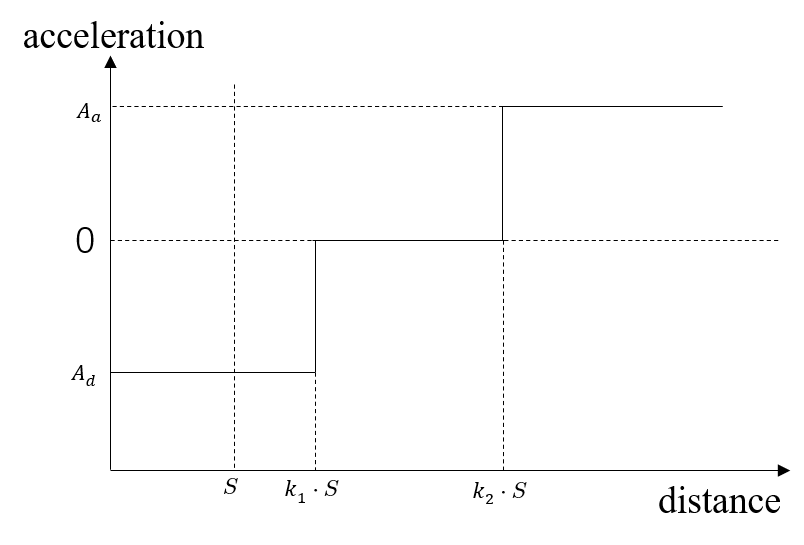}}
\caption{Safe Distance Mapping}
\label{fig:distMap}
\end{figure}

To further reduce the speed changing, we apply the safe distance mapping algorithm. If the current distance $d_{i,i+1}$ between a vehicle and its preceding vehicle satisfy $k_1\cdot S_{i,i+1}<d_{i,i+1}<k_2\cdot S_{i,i+1}$, where $k_2>f_k>1$, the acceleration of this vehicle should be zero. The acceleration-distance relationship is illustrated in Fig. \ref{fig:distMap} and can be expressed by the following equation:

\begin{equation}
\label{eq:distMap}
a_i=\left\{
\begin{array}{ll}
A_a, & d_{i,i+1}>k_2\cdot S_{i,i+1}\\
0, & k_1\cdot S_{i,i+1}<d_{i,i+1}<k_2\cdot S_{i,i+1}\\
A_d, & d_{i,i+1}<k_1\cdot S_{i,i+1}
\end{array}
\right.,
\end{equation}

\begin{figure}[htbp]
\centerline{\includegraphics[width=0.5\textwidth]{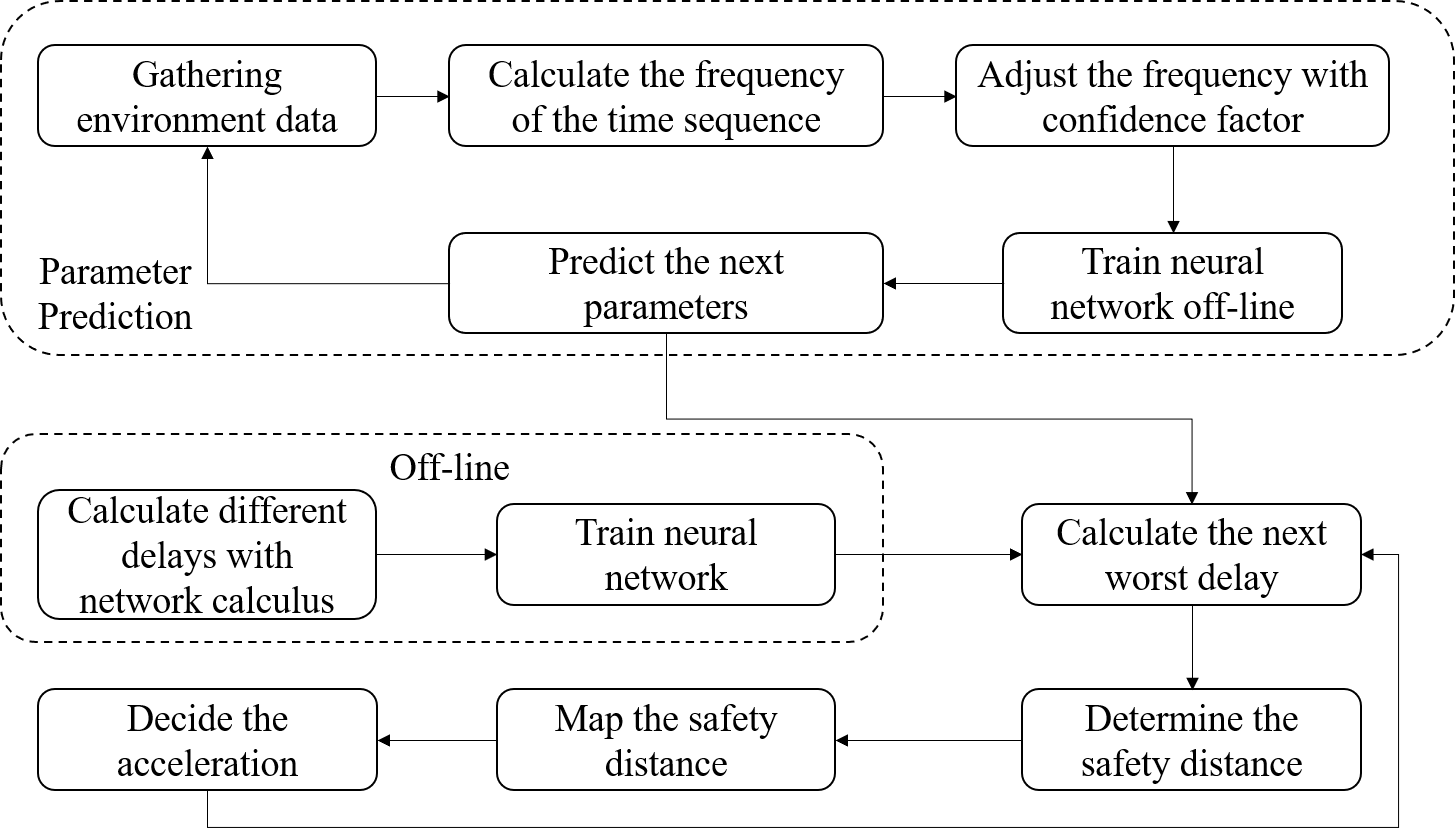}}
\caption{The flowchart of the intelligent distance control strategy.}
\label{fig:flowchart}
\end{figure}

\noindent where $A_a$ is a positive number and $A_d$ is a negative number, which can be chosen according to the traffic. Combining the parameter prediction, distance mapping and neural network calculate delay upper bound, the whole intelligent distance control strategy is given by Algorithm \ref{alg:following}. The flowchart of the whole process is presented in Fig. \ref{fig:flowchart}.

\begin{algorithm}[t] 
\caption{Intelligent distance control strategy}

\begin{algorithmic}[1]
 \STATE \textit{Initialization}: Calculate different delays with various parameters such as $p,\lambda,\xi,n$.
 \STATE \textit{Initialization}: Train a neural network with samples created from the above calculated results.
 \WHILE {True}
  \STATE Gathering environment data for the prediction. 
  \STATE Predict the noise power, number of vehicles and packet arrival rate with Algorithm \ref{alg:prediction}.
  \STATE Calculate the worst delay upper bound $D$ in the next $T_n$ time with the trained neural network.
  \STATE Calculate the safe distance $S$ with equation \eqref{eq:safetyDistance}.
  \STATE Map $S$ and decide the acceleration with equation \eqref{eq:distMap}.
  \ENDWHILE
  
 \end{algorithmic} 
\label{alg:following} 
\end{algorithm}

\section{Performance Evaluation}
\label{simulation}

\begin{figure}[htbp]
\centerline{\includegraphics[width=0.4\textwidth]{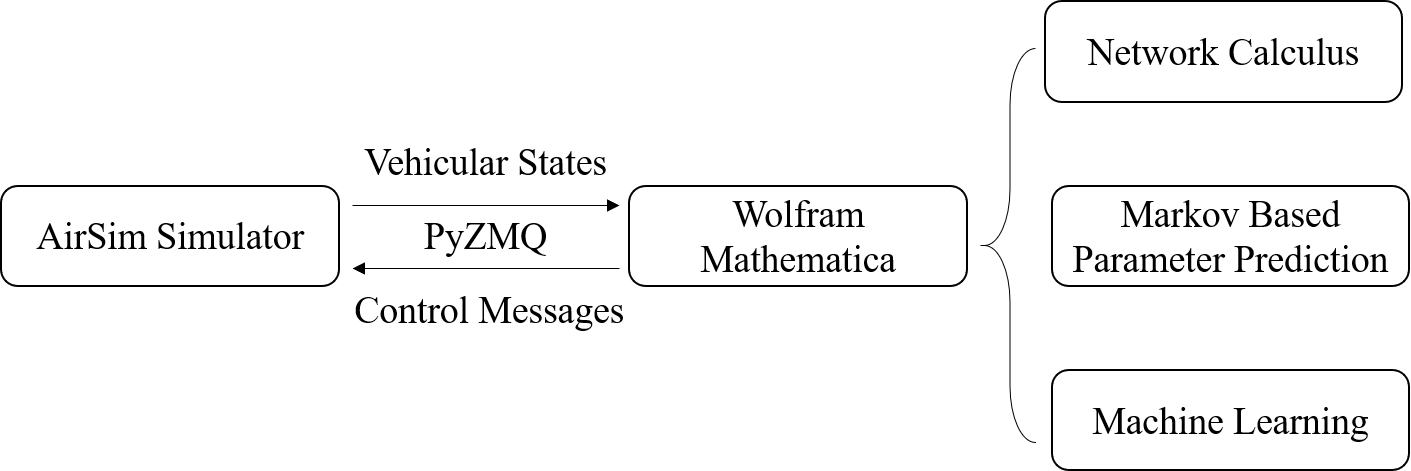}}
\caption{The simulation framework.}
\label{fig:simulation}
\end{figure}

\begin{figure}[htbp]
\centerline{\includegraphics[width=0.4\textwidth]{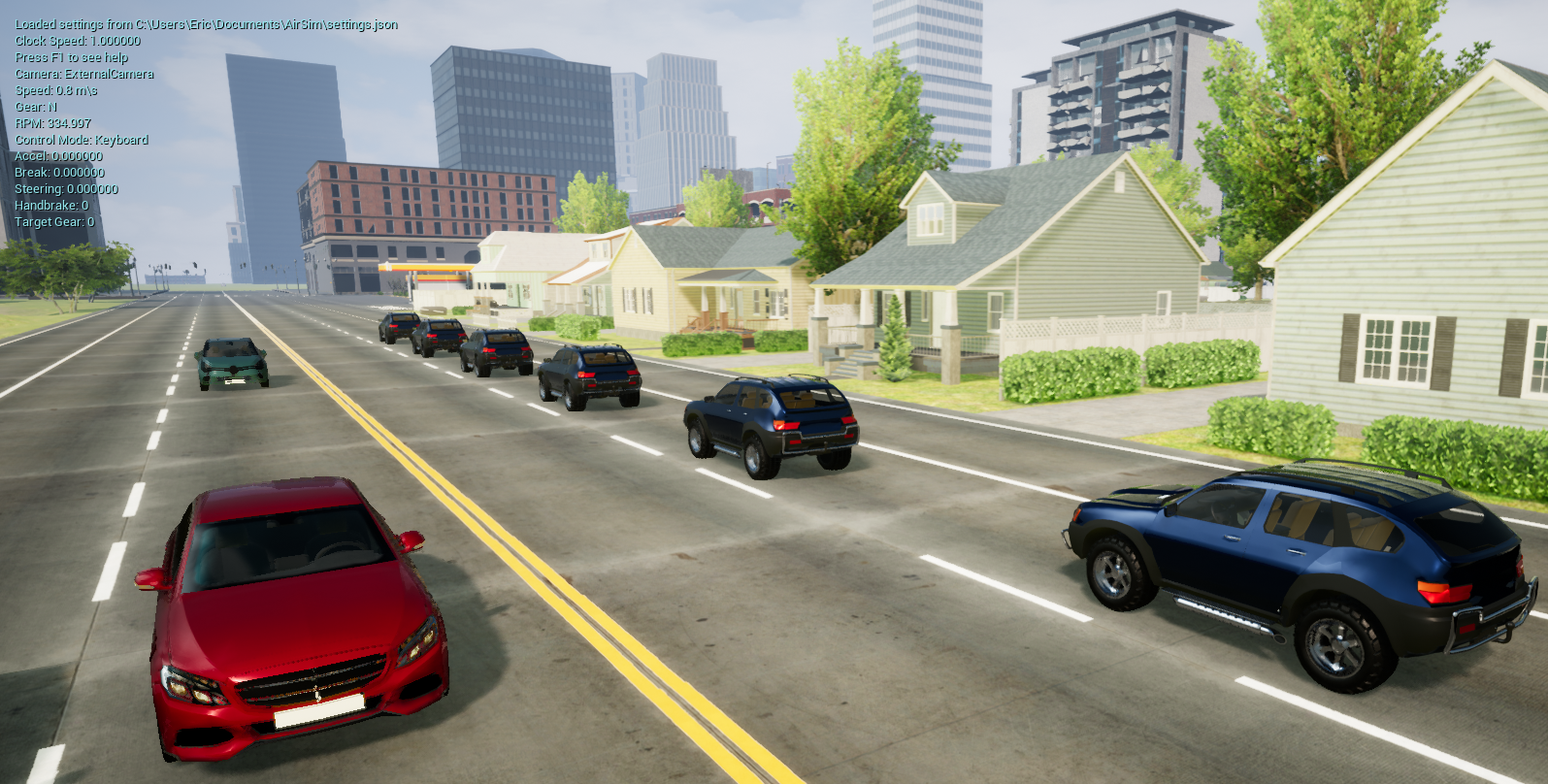}}
\caption{A screen shot of the AirSim simulator.}
\label{fig:airsim}
\end{figure}

In this section, we present the simulation results for the delay upper bounds of cellular V2V, 6G communications and hybrid V2V networks, respectively and the proposed intelligent distance control strategy. In the traffic control simulation, we use the AirSim \cite{airsim} simulator to conduct the real-time evaluation of the proposed algorithms. As shown in Fig. \ref{fig:simulation}, we implement an advanced network level simulation platform, including AirSim for autonomous driving simulation, Mathematica for SNC calculation, machine learning and communication simulation. In the autonomous driving simulation scenario shown in Fig. \ref{fig:airsim}, a vehicular platoon with 6 vehicles is running on one lane. Some key parameters are presented in Table \ref{tab:sim}.

\begin{table}[htbp]
\caption{Key Parameter Settings in Simulation}
\begin{center}
\begin{tabular}{|c|c|c|c|}
\hline
\textbf{Parameter}&\textbf{Value}&\textbf{Parameter}&\textbf{Value} \\
\hline
SNR in \eqref{eq:SNR}& 20dB & $W$ & 4\\
\hline
$m$ & 3 & $M$ & 5\\
\hline
$t_s$ & 10& $t_C$ & 5\\
\hline
$L$ & 100 & $l$ & 5m\\
\hline
\end{tabular}
\label{tab:sim}
\end{center}
\end{table}

\begin{figure}[htbp]
\centerline{\includegraphics[width=0.4\textwidth]{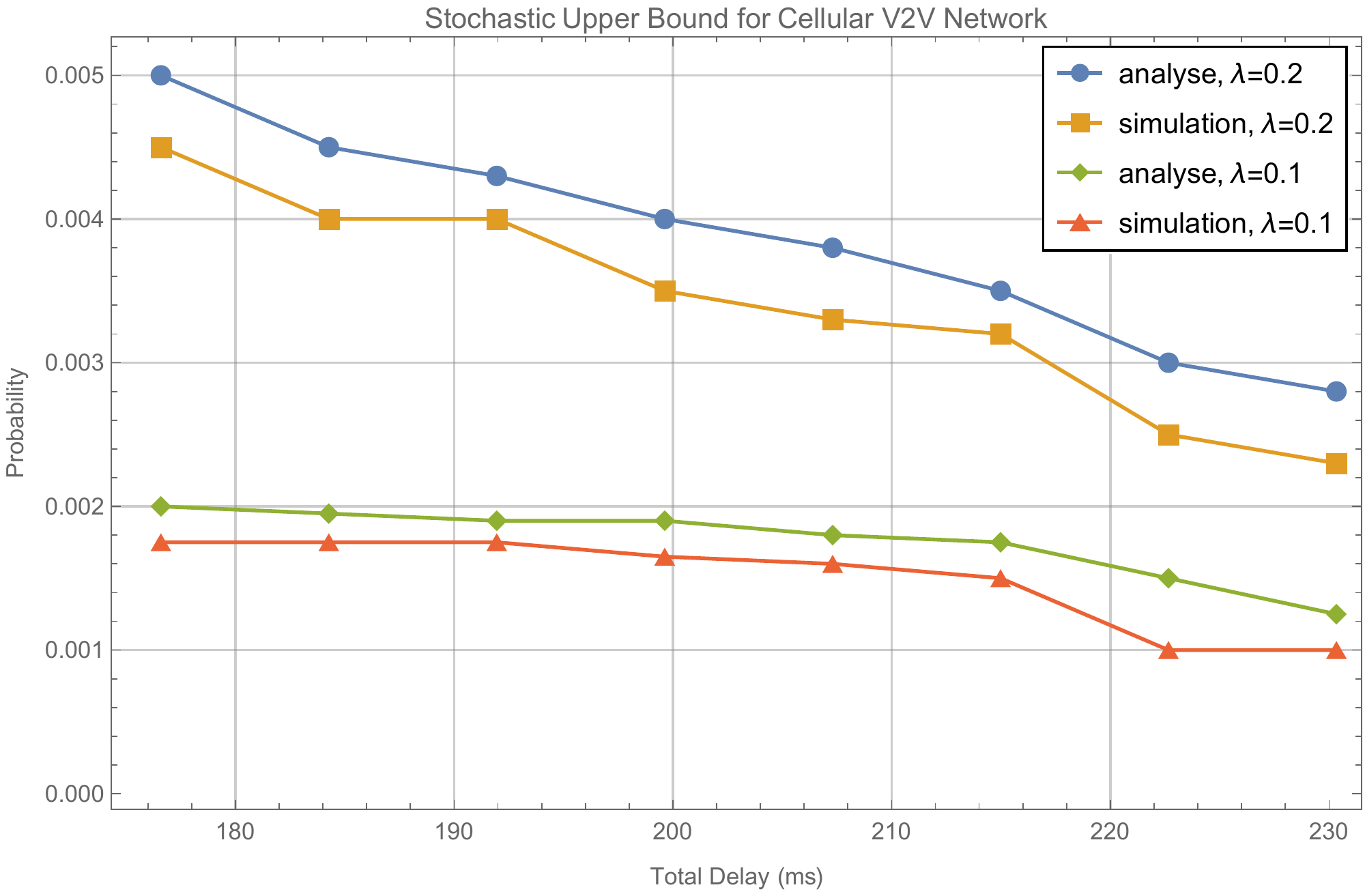}}
\caption{Stochastic delay upper bound for pure cellular V2V network.}
\label{fig:DSRC}
\end{figure}

Fig. \ref{fig:DSRC} shows the stochastic delay upper bound with pure cellular V2V communication. The number of vehicles is 10 in the simulation. The simulation delay is the probability of 10000 simulations. When the arrival rate is small ($\lambda=0.1$), the tail of the delays' Probability Density Function (PDF) keeps small and decreases in a gentle trend. However, when the arrival rate becomes high ($\lambda=0.2$), the PDF decreases fast and becomes indistinguishable from the other PDF at the tail. Moreover, Fig. \ref{fig:DSRC} also shows that the delay upper bound with cellular V2V communication with competitive access protocols is very tight.

\begin{figure}[htbp]
\centerline{\includegraphics[width=0.4\textwidth]{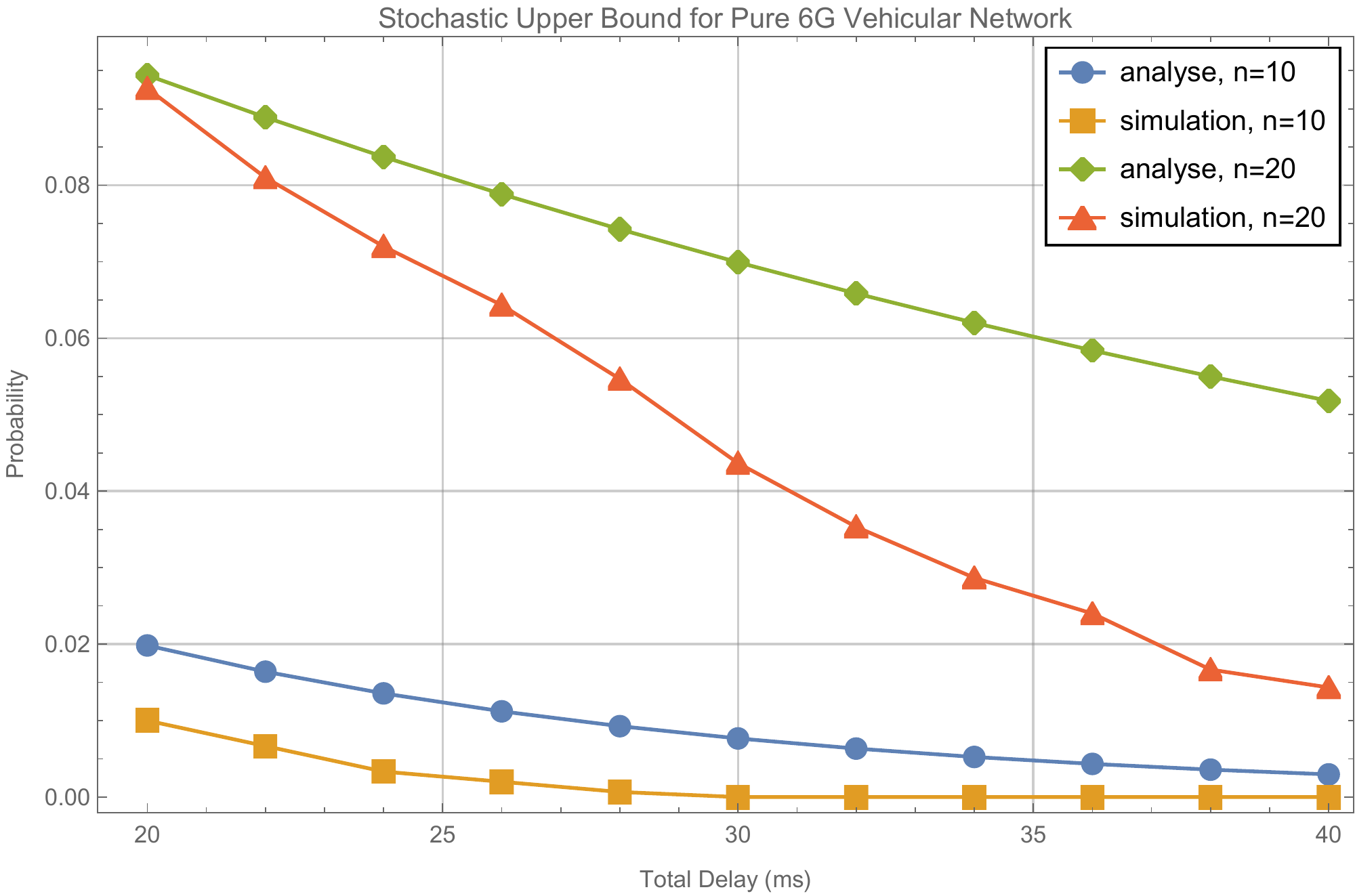}}
\caption{Stochastic delay upper bound for pure 6G communication network.}
\label{fig:mmWave}
\end{figure}

We illustrate the stochastic delay upper bound with mmWave communication in Fig. \ref{fig:mmWave} as a typical representation of 6G vehicular communication technologies, where the packet arrival rate is 0.1. The delay with mmWave is less than that with cellular V2V communication, due to the high speed of the mmWave links. However, the number of vehicles will affect the mmWave network dramatically. This is because multi-hop route increases the queue waiting time, which is the major contribution to the mmWave communication  delay. As shown in Fig. \ref{fig:mmWave}, twice of the vehicle number increase the delay about four times.

\begin{figure}[htbp]
\centerline{\includegraphics[width=0.4\textwidth]{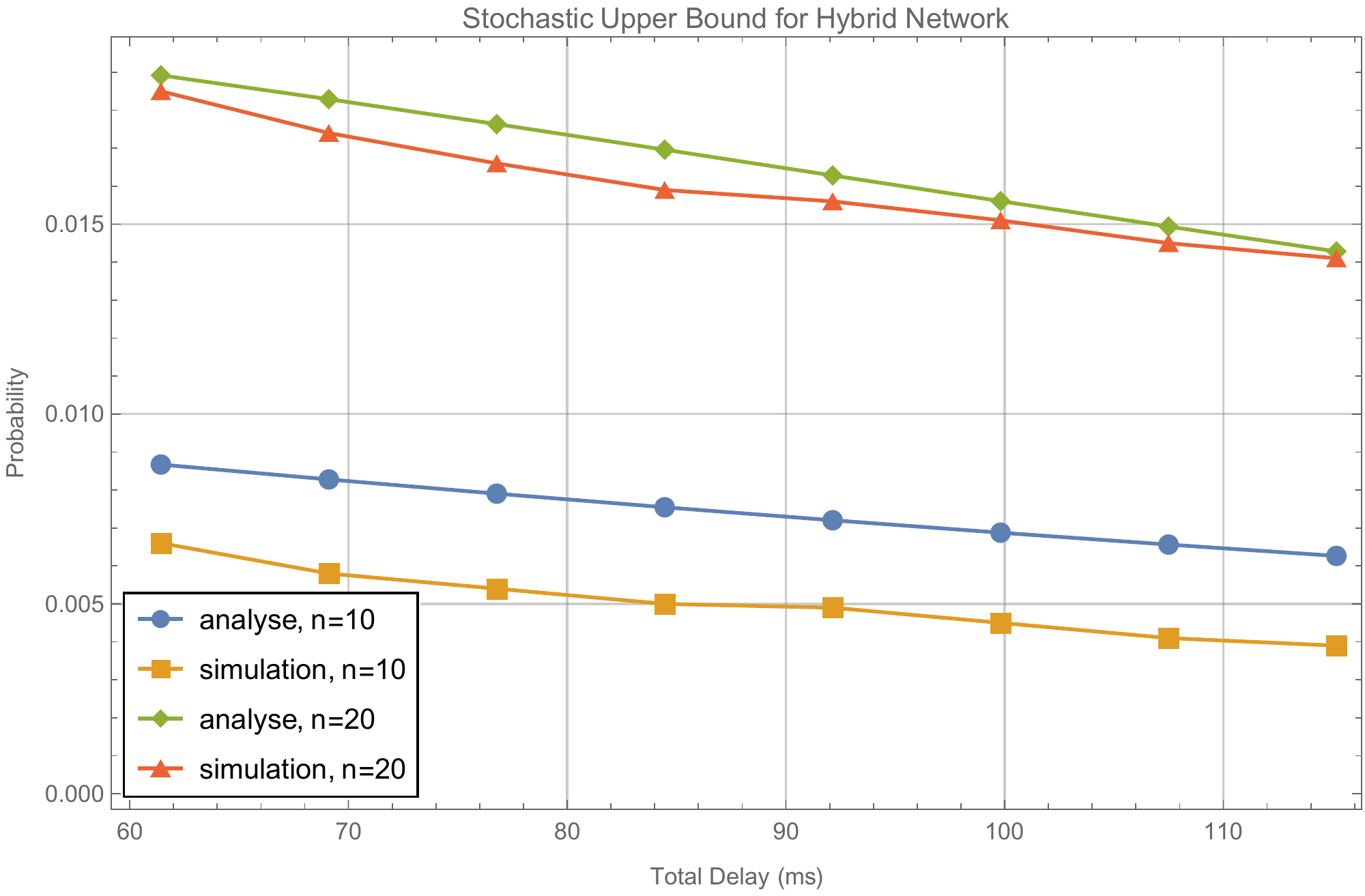}}
\caption{Stochastic delay upper bound for hybrid V2V network}
\label{fig:hybrid}
\end{figure}

In Fig. \ref{fig:hybrid}, the stochastic delay upper bound for hybrid V2V network is presented where packet arrival rate is 0.2 and the traffic is equally divided to be transmitted with mmWave and cellular V2V communications. Compare to that of cellular V2V communication, the tail probability with both communication technologies is smaller, which means a faster communication speed. This is because of the contribution of the mmWave communication. Moreover, compared with mmWave communication, the impact of the number of vehicles becomes lighter, which is caused by the one-hop communication pattern. As a result, the network with both communication technologies has better performance in most cases than the homogeneous networks.

\begin{figure}[htbp]
\centerline{\includegraphics[width=0.4\textwidth]{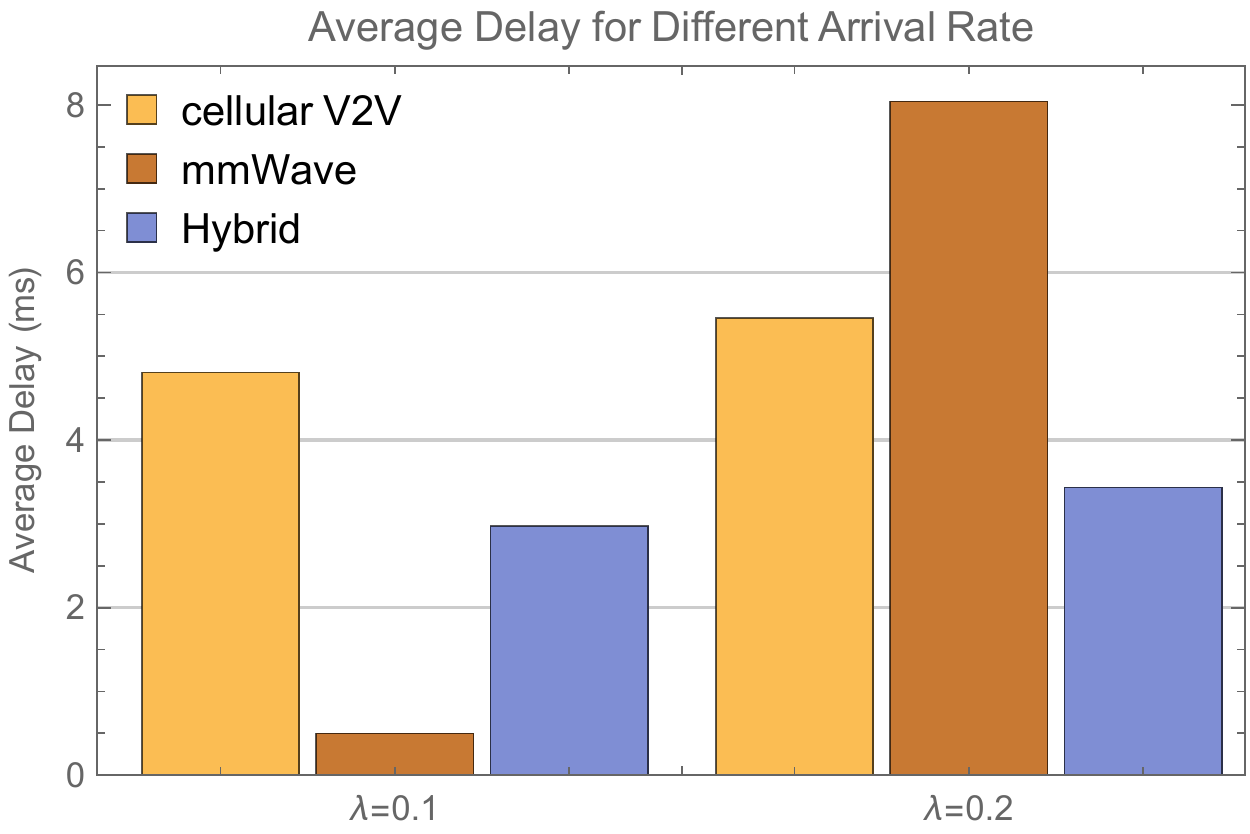}}
\caption{Average delay for different arrival rates.}
\label{fig:average}
\end{figure}

Fig \ref{fig:average} shows the average delay  versus different packet arrival rate. When the packet arrival rate is small, the mmWave communication delivers  the best performance. However, when the arrival rate increases, the performance with mmWave communication drops dramatically. This is because that each vehicle generates packets in the same packet arrival rate, which makes the queue long in a high speed, which reduces the performance of the mmWave communication. And the hybrid network can offload some of the traffic to the cellular V2V communication channel to keep the network in a normal state. Consequently, the hybrid  network provides better performance guarantee than the vehicular networks with only mmWave communications.

\begin{figure}[htbp]
\centerline{\includegraphics[width=0.4\textwidth]{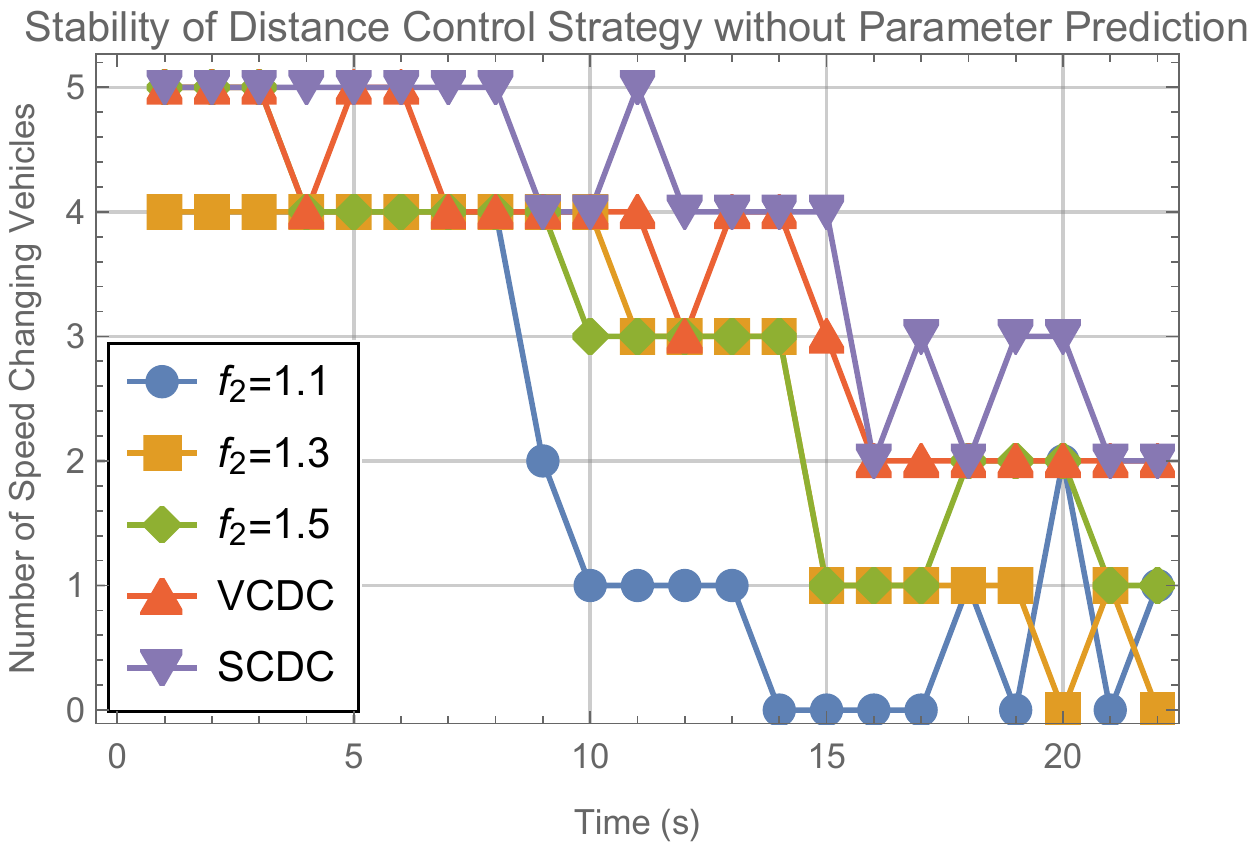}}
\caption{Stability of distance control strategy}
\label{fig:stability}
\end{figure}

We illustrate the stability of the intelligent distance control strategy in Fig. \ref{fig:stability}. We consider the algorithms with smoother vehicular speed changing and higher converge speed have higher stability. We compare the vehicular control algorithms only based on the sensor data in the traffic control fields, which is called String Constrained Distributed Control (SCDC)\cite{SCDC} and Velocity Constrained Distributed Control (VCDC)\cite{VCDC}, respectively. Compared to the algorithms without communication, the proposed algorithms converge much faster. At the beginning, there are 6 vehicles in the simulator and the distance between the adjacent vehicles is 10 meters and the speed of the vehicular is 7.5 meters per second. Because the communication condition becomes bad, the safe distance becomes longer and vehicles need to change their relative positions. The new safe distance is about 11.5 meters. Therefore, all of the vehicles beside the first vehicle need to decelerate. We count the number of vehicles that change their speeds at the different time slot. Lower $f_2$ means shorter safe distance mapping in equation \eqref{eq:distMap}. As a result, the vehicular platoon becomes stable faster if the $f_2$ is lower. However, all of the vehicular platoon become stable after 20 seconds with flat speed changing.

\begin{figure}[htbp]
\centerline{\includegraphics[width=0.4\textwidth]{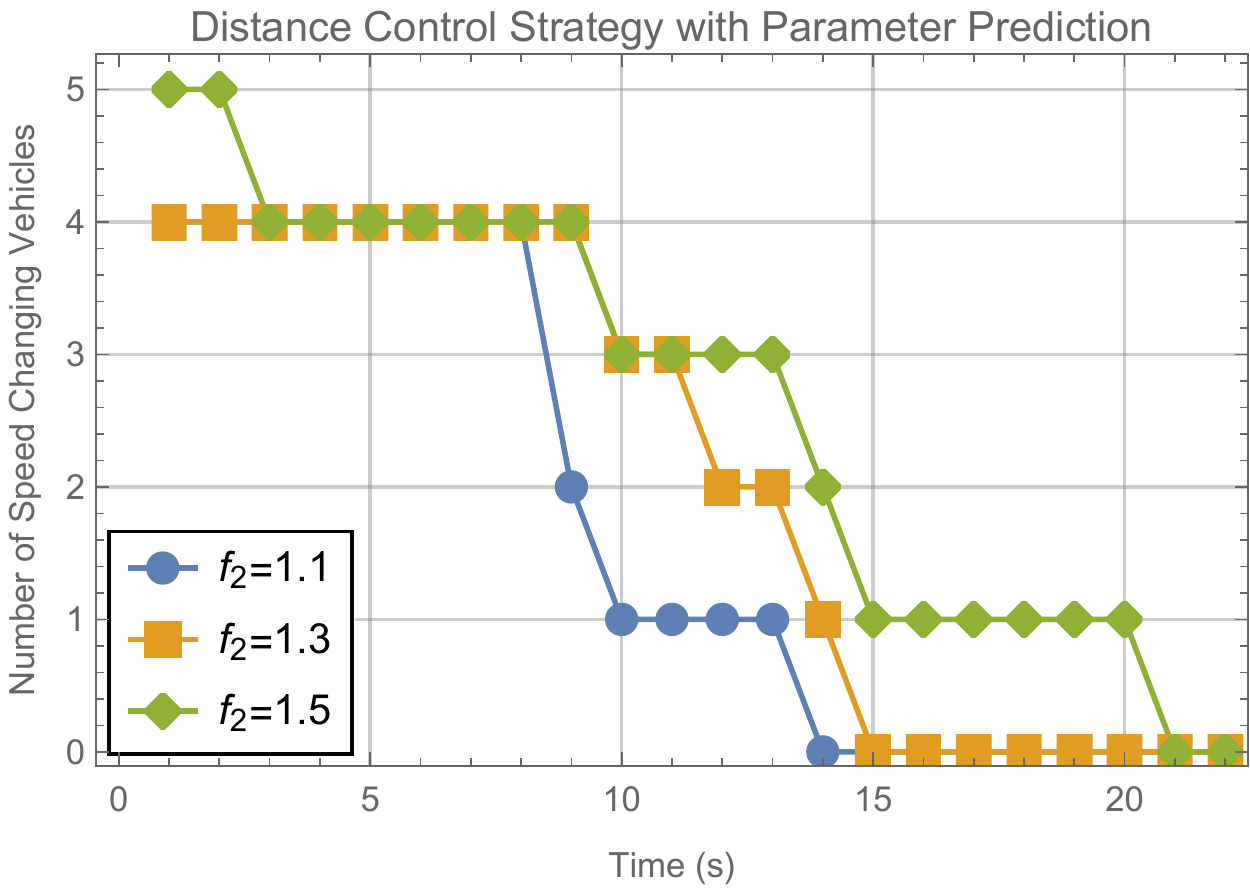}}
\caption{Distance control strategy without parameter prediction}
\label{fig:predict}
\end{figure}

Fig. \ref{fig:predict} illustrates the distance control strategy with parameter prediction in the same scenario with Fig. \ref{fig:stability}. Because all the vehicles decide their speeds only according to the current environment in the previous scenario, without parameter prediction, the vehicular platoon still fluctuates after becoming stable. With the parameter prediction algorithm, the number of speed changing vehicles becomes much less in the stable period.

\begin{figure}[htbp]
\centerline{\includegraphics[width=0.4\textwidth]{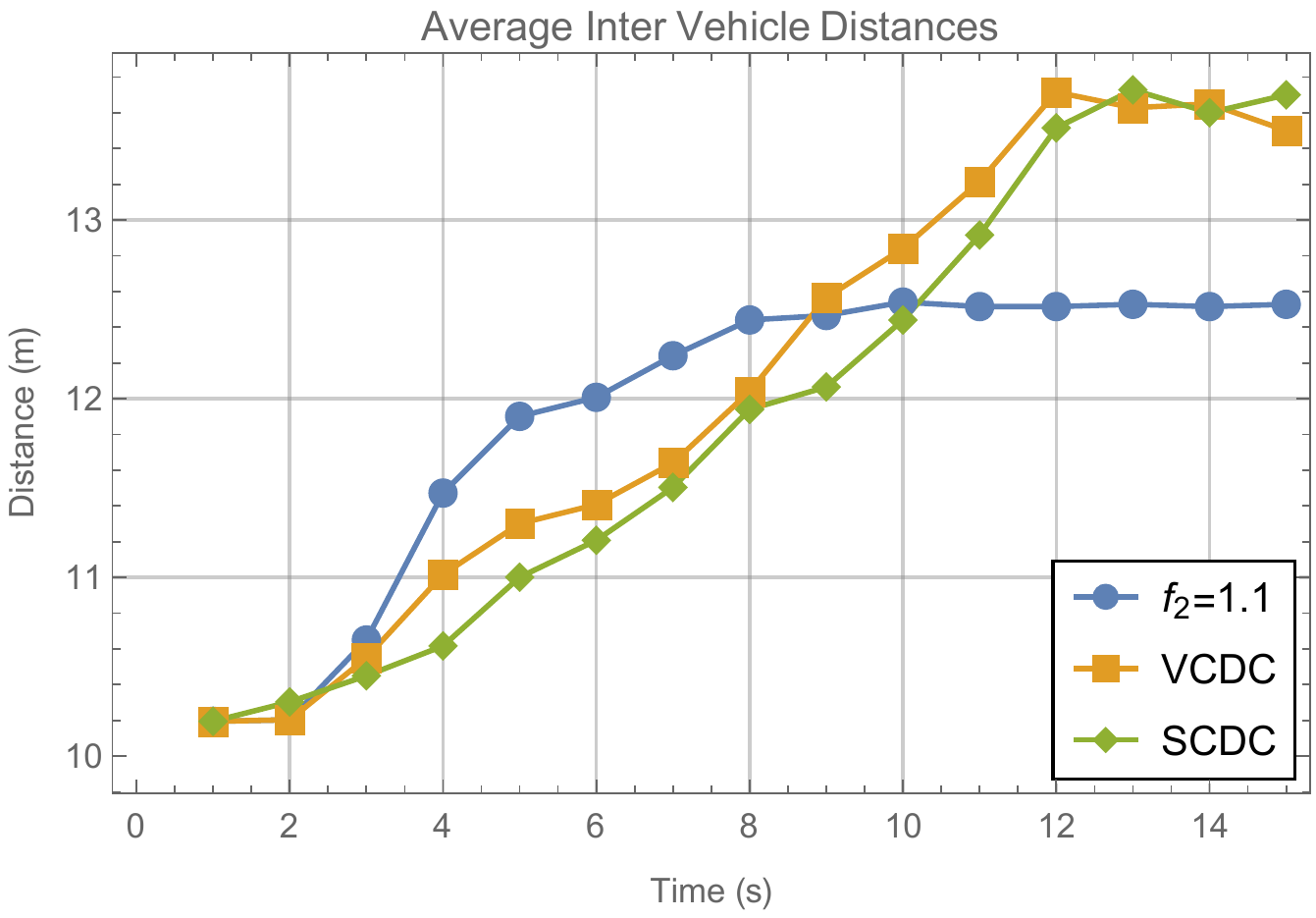}}
\caption{Distances between vehicles in intelligent distance control strategy.}
\label{fig:distances}
\end{figure}

Fig. \ref{fig:distances} presents the average inter vehicle distance of different algorithms. The simulation parameters are the same as the simulation in Fig. \ref{fig:stability} where $f_2=1.1$. The distances become larger slowly and converge after 10 seconds. After the platoon becomes stable, the distances fluctuate around the stable distance according to the delay upper bound. Compared with the other control algorithms, the intelligent distance control converges faster and becomes more stable at the end of simulation. In addition, the average inter-vehicle distance is much shorter than other algorithms, which means higher traffic efficiency than individual autonomous driving methods. Furthermore, according the the traffic laws in China, the safety distance should be larger than 30 meters, which means the 6G supported cooperative autonomous driving with proposed distance control strategy has 3 times road capacity compare with manual driving methods, which is a dramatic improvement for road efficiency.

\begin{figure}[htbp]
\centerline{\includegraphics[width=0.4\textwidth]{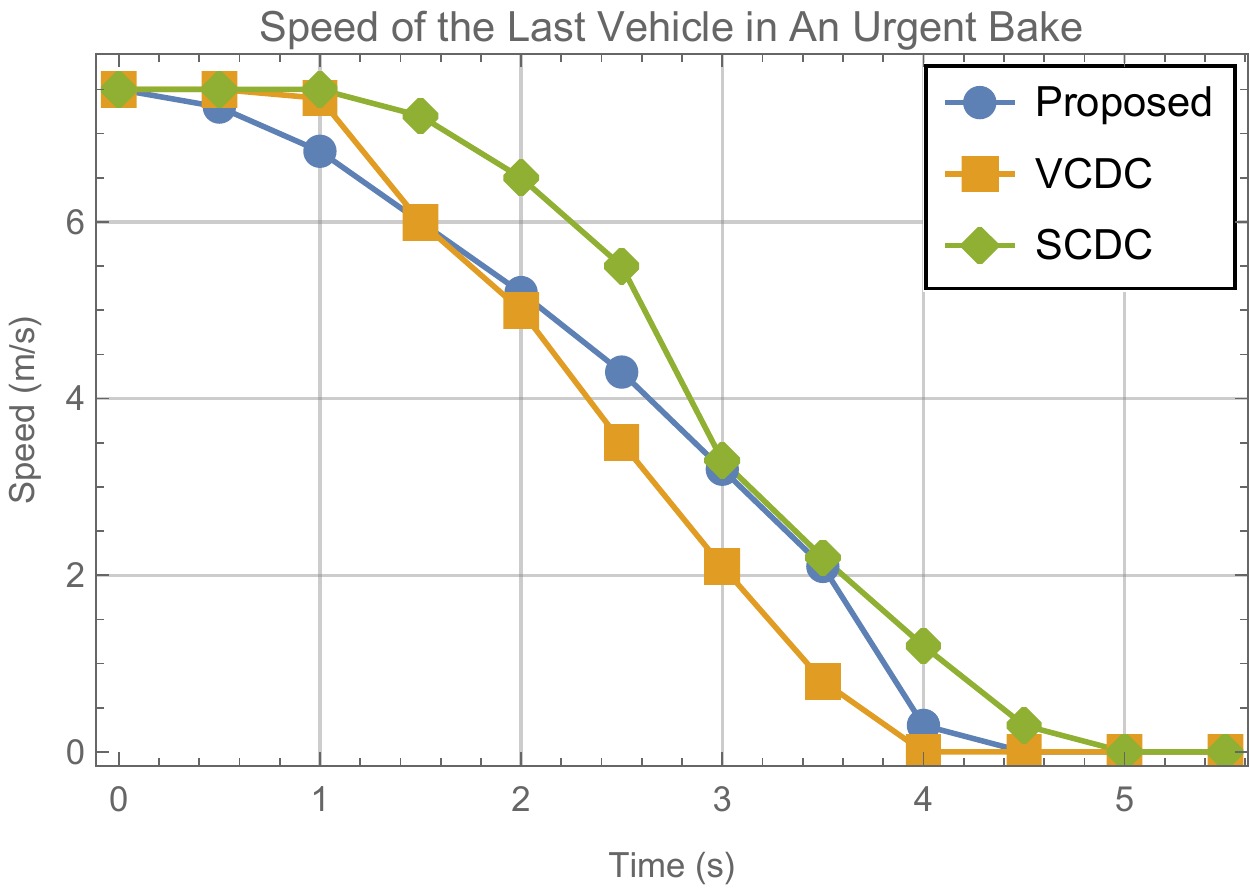}}
\caption{Speed of the last vehicle in an urgent bake.}
\label{fig:safety}
\end{figure}

Furthermore, we conduct the simulation experiment for an urgent bake scenario. The first vehicle in the platoon start to bake and send control messages through vehicular network. The speed of the last vehicle in the platoon is recorded. From Fig. \ref{fig:safety}, we can find the last vehicle starts to brake in millisecond level, which is much faster than the reaction speed in the individual autonomous driving modes. This is because that the individual vehicles need to observe the behaviours of its preceding vehicle to decide its speed, which accumulate lots of reaction times along the lined platoon. Consequently, communication based vehicle control can enhance safety because of the fast reaction for the accidents. Moreover, the smooth speed changing due to the proposed distance control algorithm can provide better driving experience compared to other control algorithms.

The above simulation results show that the network calculus can accurately model the delay upper bound. We can predict the delays precisely based on the deep understanding of the network states. Using the intelligent distance control algorithm, the platoon driving can become stable with smooth speed changing.

\section{Conclusion}
\label{conclusion}

In this paper, we study the communication performance bounds for 6G vehicle networks with hybrid communication and channel access schemes.
Based on the proposed stochastic network calculus analytical model, we derive the performance of the cellular V2V communication networks and high rate 6G communication networks.
Then, the delay low bound of V2V communications is obtained by using Stieltjes convolution. 
A deep learning based approach is designed to predict the V2V communication
performance bounds for real time operation. 
Based on the performance guarantee analytical model, we propose
an intelligent inter-vehicle distance control strategy for cooperative driving. Moreover, a new parameter prediction algorithm is designed to reduce
the number of speed changes for smooth driving. 
Extensive simulation experiments are conducted by an integrated simulation platform that combines communication module, analytical
module and AirSim driving module. Simulation results show that the delay upper bound analysis and parameter prediction
have sufficient accuracy for practical use. The proposed algorithms are very effective in maintaining inter-vehicle safe distance
and stabilizing the cooperative driving platoons, which is helpful to improve driving safety, traffic and energy efficiency.
As the simulation results show, the proposed algorithm can improve the road capacity about 10\% than the existing individual autonomous driving based safe distance control methods.

\bibliographystyle{IEEEtran}
\bibliography{bibs}

\end{document}